\documentclass[10pt,conference]{IEEEtran}
\IEEEoverridecommandlockouts
\usepackage{cite}
\usepackage{amsmath,amssymb,amsfonts}
\usepackage{algorithmic}
\usepackage{graphicx}
\usepackage{textcomp}
\usepackage{hyperref}
\usepackage{xcolor}
\usepackage{url}
\usepackage[inline]{enumitem}
\def\BibTeX{{\rm B\kern-.05em{\sc i\kern-.025em b}\kern-.08em
    T\kern-.1667em\lower.7ex\hbox{E}\kern-.125emX}}
\begin{document}

\IEEEaftertitletext{\vspace{-3\baselineskip}}

\title{YABLoCo: Yet Another Benchmark for Long Context Code Generation}

\author{
    \IEEEauthorblockN{Aidar Valeev}
    \IEEEauthorblockA{
    \textit{Research Center of the Artificial Intelligence Institute} \\
    \textit{Innopolis University, Russia} \\
    ai.valeev@innopolis.ru}
    \and
    \IEEEauthorblockN{Roman Garaev}
    \IEEEauthorblockA{
    \textit{Research Center of the Artificial Intelligence Institute} \\
    \textit{Innopolis University, Russia} \\
    o.garaev@innopolis.university}

    \and
    \IEEEauthorblockN{Vadim Lomshakov}
    \IEEEauthorblockA{
    \textit{St. Petersburg Department of the Steklov Institute of Mathematics, Russia} \\
    vadim.lomshakov@gmail.com}
    \and
    \IEEEauthorblockN{Irina Piontkovskaya}
    \IEEEauthorblockA{
    \textit{Huawei Noah's Ark Lab} \\
    piontkovskaya.irina@huawei.com}
    
    \and
    \IEEEauthorblockN{Vladimir Ivanov}
    \IEEEauthorblockA{
    \textit{Research Center of the Artificial Intelligence Institute} \\
    \textit{Innopolis University, Russia} \\
    v.ivanov@innopolis.ru}
    \and
    \IEEEauthorblockN{Israel Adewuyi}
    \IEEEauthorblockA{
    \textit{Research Center of the Artificial Intelligence Institute} \\
    \textit{Innopolis University, Russia} \\
    i.adewuyi@innopolis.university}
}


\maketitle

\begin{abstract}
Large Language Models (LLMs) demonstrate the ability to solve various programming tasks, including code generation. Typically, the performance of LLMs is measured on benchmarks with small or medium-sized context windows of thousands of lines of code (LoC). At the same time, in real-world software projects, repositories can span up to millions of LoC. 
This paper closes this gap by contributing to the long context code generation benchmark (YABLoCo). The benchmark featured a test set of 215 functions selected from four large repositories with thousands of functions. The dataset contained metadata of functions, contexts of the functions with different levels of dependencies, docstrings, functions' bodies, and call graphs for each repository. This paper presents three key aspects of the contribution. 
First, the benchmark aims at function body generation in large repositories in C and C++, two languages not covered by previous benchmarks. Second, the benchmark contains large repositories from 200K to 2,000K LoC.  
Third, we contribute a scalable evaluation pipeline for efficient computing of the target metrics and a tool for visual analysis of generated code. Overall, these three aspects allow for evaluating code generation in large repositories in C/C++. The dataset as well as the code for the evaluation pipeline are available  at 
\url{https://github.com/yabloco-codegen/yabloco-benchmark}

\end{abstract}

\begin{IEEEkeywords}
Large Language Models for Code, Code Generation Benchmarks, Large Repository Context, C/C++.
\end{IEEEkeywords}


\section{Introduction}
\vspace*{-0.1cm}
Large Language Models (LLMs) have recently demonstrated abilities to solve a wide set of software engineering tasks in various settings \cite{jiang2024survey, zhang2023survey}.
Code generation tasks typically include code autocompletion of up to 1-3 LoC and code generation where an LLM should generate a function body or even a class\cite{classeval}. Many LLMs for code position themselves as code assistants. They could be split into two groups: chat-assistants answering questions on code and provided context, such as CodeLlama\cite{CodeLlama}, DeepSeek Coder\cite{Deepseekcoder}, and Codestral\footnote{\url{https://mistral.ai/news/codestral/}}, and the assistants built-in into IDE that propose code completion considering whole repository context, such as GitHub Copilot\footnote{\url{https://github.com/features/copilot/}} and Codeium \footnote{\url{https://codeium.com/}}). With both groups, the context plays a crucial role, as well as the prompt. Their performance is measured on small or medium-sized context windows up to thousands of LoC. Even in these setups, not all functions generated by them compile and pass tests.

In program synthesis, a model should generate a function in a specific programming language, given a text description of a function. A straightforward approach is prompting an LLM with a function's docstring and its signature, assuming that the docstring describes functionality. Typically, the performance of LLMs for code generation is measured on several public benchmarks, such as HumanEval\cite{HumanEval} and MBPP EvalPlus\footnote{\url{https://evalplus.github.io/}}, that allow to calculate the pass@k metric. However, these benchmarks only cover simple independent functions, while functions in the real world are interdependent pieces of complex projects. 

In the software engineering industry, real-world repositories can span up to millions of LoC. In such cases, the situation with evaluation changes dramatically. Modern LLMs are struggling with extracting and analyzing large contexts let alone the code generation. Clearly, extracting a key or statement from a long context is simpler than understanding all inter-function dependencies in a long repository. 
The state-of-the-art models claim to have context length up to 1M+ tokens\footnote{\url{https://github.com/hsiehjackson/RULER}}.
This scale leaves much room to evaluate and improve code generation, involving such challenges as cross-repository context, cross-language modality, cross-file generation, test generation, and many others. However, lacking an adequate benchmark is an obstacle in this direction.

Several code generation studies propose repository-level benchmarks \cite{CoderEval,RepoBench,CrossCodeEval,RepoFusion, RepoCoder,pan2024enhancing,Li2024EvoCodeBenchAE}. These benchmarks differ in three dimensions. Firstly, some models generate whole functions, while others generate statements following the Fill-in-the-Middle objective. Secondly, many models evaluate results by running tests and computing the pass@k metric, while others only calculate the `Exact Match' and the `Edit Similarity' metrics. Finally, the existing benchmarks focus mostly on Python and Java programming languages, and none on C/C++. Recently, the Long Code Arena set of benchmarks has been introduced \cite{bogomolov2024long}, but this set also focuses on Python only.
Yet, no benchmark and evaluation tools are available for C/C++ to adequately assess LLM performance in the large repository context setting. 

This paper closes this gap by contributing to the long context code generation benchmark (YABLoCo). The benchmark contains 215 examples of functions along with function bodies, docstrings, signatures, and other metadata. The benchmark can be used to evaluate the effectiveness of code generation models while capturing the relevant context in large repositories and generating runnable code. YABLoCo benchmark balances different context types and allows for functions with dependencies at different levels, including the same file, as well as dependencies from other files and packages. 

Specifically, this paper focuses on the following three aspects:
\begin{enumerate}
    \item a dataset for evaluating code generation on C/C++ programming languages in large repositories containing hundreds of thousands of LoC;
    \item an open-source pipeline for benchmark collection, evaluation, and visual analysis of generated code;
    \item the quality evaluation of SoTA LLMs for code generation models in different settings.
\end{enumerate}



\section{Benchmark collection}
\label{sect:Bench}
This section describes the approach and results of data collection that includes repository selection, pre-processing and filtering, metadata extraction, selection of functions with docstrings, test coverage analysis, and analysis of the resulting dataset quality. This data collection was inspired by the CoderEval \cite{CoderEval}, a benchmark with different levels of context for Python and Java. CoderEval contains a similar number of examples, 230 examples from 10 projects in Java and 230 examples from 43 projects in Python. Hence, some of our decisions are justified by their methodology.

\subsection{Repository selection and preprocessing}
A critical part of long context code benchmark creation is selecting repositories properly. A properly-selected repository should aim at hundreds of thousands of lines of C and C++ code. 
In addition, a repository should satisfy minimum code style standards. The latter requirement is related to the popularity of a repository: if it is one of the most starred on GitHub, we assumed that its code was readable and understandable enough. 

The resulted list is $llvm\mbox{-}project$, $bullet3$, $openssl$, and $redis$. The $llvm\mbox{-}project$ \footnote{\url{https://github.com/llvm/llvm-project}} is a C/C++ repository of a toolkit for compilers and run-time environments. The $bullet3$ \footnote{\url{https://github.com/bulletphysics/bullet3}} is a C++ repository of a software development kit for bullet physics engine. The $openssl$ \footnote{\url{https://github.com/openssl/openssl}} is a C repository of an open-source toolkit for security protocols. Finally, the $redis$ \footnote{\url{https://github.com/redis/redis}} is a C repository of a NoSQL database server. We rejected certain most starred candidate repositories, such as $linux$, due to their extraordinary complexity. Note that, two of the selected repositories are in the Stack dataset that is typical pre-training data for LLMs. This may lead to potential data leakages; we address this issue in Section \ref{sec:limitations} below.

Table \ref{tab:cloc} below presents descriptive statistics for the number of files as well as the number of lines of code computed with the $cloc$ tool\footnote{\url{https://github.com/AlDanial/cloc/releases/tag/v1.90}} for each repository.

\begin{table}[ht]
    \vspace{-5pt}
    \centering
    \caption{\textit{cloc} utility statistics for the selected repositories}
    \begin{tabular}{|l|p{0.58cm}p{0.58cm}|p{0.58cm}p{0.58cm}|p{0.58cm}p{0.58cm}|p{0.58cm}p{0.58cm}|}
    \hline
         &  \multicolumn{2}{c|}{C++}  &  \multicolumn{2}{c|}{C}  &  \multicolumn{2}{c|}{C/C++ Header}  &  \multicolumn{2}{c|}{Overall}\\
         Project & Files & Lines & Files & Lines & Files & Lines & Files & Lines \\
    \hline
         llvm & 30.2K & 5.6M & 10.7K & 1M & 12.1K & 1.5M & 120K & 29.6M\\
         bullet3 & 751 & 751K & 67 & 45.3K & 1426 & 590K & 4269 & 1.9M\\
         openssl & - & - & 1554 & 530K & 506 & 65.6K & 2903 & 1.1M\\
         redis & 7 & 402 & 421 & 173K & 279 & 27.9K & 1497 & 428K\\
    \hline
    \end{tabular}
    \vspace{-10pt}
    
    \label{tab:cloc}
\end{table}

\subsection{Context-based categorization of functions}
From each of the selected repositories, we extracted all functions along with their function calls, last commit date, docstring comment, code and comment length, and test hits. To this end, a clang-based tool was used to build and analyze function call graphs, one graph per repository. Table \ref{tab:graph_info} below presents the overall number of functions and calls.
Each function call was represented as an edge in the call graph. The function calls were then assigned to one of the
following five categories: \\
\begin{itemize*}
    \item `\textit{none}'
    \item `\textit{stdlib}'
    \item `\textit{file}'
    \item `\textit{package}'
    \item `\textit{project}'.
\end{itemize*}

\begin{table}[]
    \vspace{-15pt}
    \centering
    \caption{Statistics over whole repositories that form the benchmark}
    \begin{tabular}{|l|r|r|} \hline
        Repository &  \# of functions (nodes) & \# of calls (edges) \\ \hline
        llvm & 175,001 & 571,883 \\
        bullet3 & 2,778 & 4,917 \\
        openssl & 19,690 & 61,998 \\
        redis & 7,905 & 24,226 \\ \hline
    \end{tabular}
    \label{tab:graph_info}
\end{table}

Specifically, `\textit{stdlib}' for system calls, `\textit{file}' and `\textit{package}' for calls inside the same file and package, correspondingly, and `\textit{project}' for functions with project-level calls. A function may have calls of several levels; in this case, we categorized it into the group of the outermost call. However, all calls of the standard library in the $bullet3$ repository accompanied another outer call, making the "stdlib" group empty. If a function had no dependencies, it went into the `\textit{none}' category. 

\subsection{Filtering of functions}
\begin{figure}
    \vspace{-5pt}
    \centering
\includegraphics[width=0.55\textwidth]{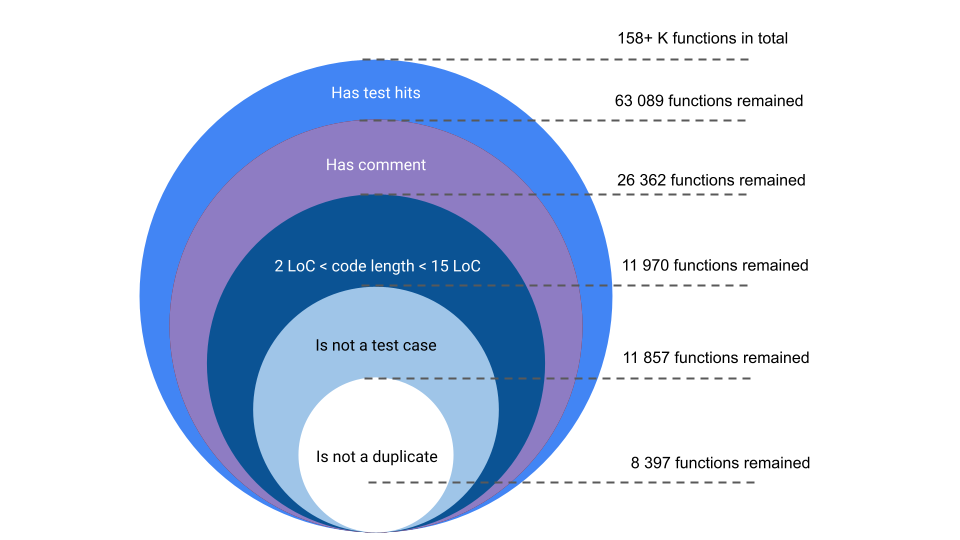}
    \caption{The counts of functions filtered at every stage}
    \label{fig:filter}
    \vspace{-15pt}
\end{figure}

We filtered, sorted and picked the functions as follows. We filtered out the functions that had excessively short or long implementations, or no test hits or comments. The counts of filtered functions are given in Fig. \ref{fig:filter}. 

We defined a proper function length between 2 and 15 lines of code. Then, we detected and removed near code duplicates. After that, we sorted the remaining set of functions in every context category according to the last commit date and test hits preferring the latest and most covered. Finally, we picked the first 12 functions from these sorted lists per context level in every repository. This automatic process produced a sample of 240 functions. Apart from the `bullet3' repository, the dataset had a balanced distribution of functions among the five groups (Table \ref{tab:bench}). The test coverage of the functions was measured with gcov tool; for each function, the number of test hits was stored in the benchmark metadata. The median test hits per function were 1,375. 

\subsection{Manual evaluation of the docstrings} \label{man_eval}
The repositories functions were sampled automatically disregarding the docstring quality. Therefore, we manually evaluated the docstring quality in the following way. Three programmers were asked to independently evaluate each function's docstring according to the question: \textit{`If this comment enough to generate the function's body?'} with three possible answers: (-)`No, the comment is not suitable', (?)`Yes, the comment is suitable, but more context is required', (+)`Yes, the comment is suitable and enough to generate function body'.
Based on their evaluations, we filtered out 25 functions with two and three negative marks from the original sample of 240 functions. 
The manual assessment results are available in the repository. 






Table \ref{table:examples_assessment} shows examples of some docstrings along with their assessments. As expected, several functions had two or three `?' marks which can be interpreted as a dependency of that function on external context. Therefore, functions with such annotations were preserved, while samples without major agreement of reviewers were dropped from the benchmark.

\begin{table}
    \centering
    \caption{Examples of docstrings with extremely negative, positive, and `questionable' marks}
    \begin{tabular}{|p{8cm}|}
     \hline
          \textbf{Negative assessment (three `-')}  \\ \hline
               Moves to the next position. \\
              \hline
              Our hash table capability is a power of two \\
              \hline
              Because we may need to allocate huge memory chunk at once when dict
resizes, we will check this allocation is allowed or not if the dict
type has resizeAllowed member function.  \\
              \hline
 \hline
    \textbf{Questionable assessment}  \\            
\hline
Find a register from its Record def. \\ \hline
 Set peer sigalg based key type \\ \hline
 Free the kvs\_it returned by kvstoreIteratorInit. \\
\hline
\hline
 \textbf{Positive assessment (three `+')}   \\ \hline
 Check if the end of a string matches 'end'  \\ \hline
Returns next non-empty dict index strictly after given one, or -1 if provided didx is the last one.   \\ \hline
 If C is a uniform value where all bits are the same (either all zero, all
ones, all undef or all poison), return the corresponding uniform value in
the new type. If the value is not uniform or the result cannot be
represented, return null.  \\
         \hline   
    \end{tabular}
    \label{table:examples_assessment}
    \vspace{-10pt}
\end{table}


\begin{table}[htbp]
    \centering
    \caption{The distribution of functions by context level group}
    \begin{tabular}{|c|c|c|c|c|c|c|}
    \hline
         \textbf{Repo Name} & \textbf{none} & \textbf{stdlib} & \textbf{file} & \textbf{package} & \textbf{project} & \textbf{Overall} \\
    \hline
         llvm & 12 & 12 & 12 & 12 & 12 & 60\\
         bullet3 & 12 & 0 & 10 & 12 & 2 & 36\\
         openssl& 12 & 11 & 12 & 12 & 12 & 59\\
         redis & 12 & 12 & 12 & 12 & 12 & 60\\
    \hline
        Total & 48 & 35 & 46 & 48 & 38 & 215\\
    \hline
    \end{tabular}
    \vspace{-10pt}
    \label{tab:bench}
\end{table}

In addition to the data collection and cleaning, we generated a call graph for each repository. The graph contained all functions with unique IDs, their callers, and callee functions as well as metadata such as length, path to file, position in file, docstring, date of the last modification,  number of test hits, and a category. 
To guarantee reproducibility, an efficient and robust evaluation pipeline is required. Section \ref{sectione} describes the implementation of the evaluation pipeline and metrics used to evaluate code generation.


\subsection{Evaluation Pipeline and Metrics}
\label{sectione}

Given an LLM that generates code, one can apply it on the benchmark. A direct approach to achieve this is `In-Context Learning' (ICL)\footnote{Here `in-context learning' term is a synonym to the `zero-shot', as we use the prompt without any context as shown in the first prompt template in Section \ref{sec:exp-setup}.}. The procedure is based on a prompt and can be used for each benchmark sample as follows.
\begin{enumerate}
    \item Combine a prompt for the sample function with name \textbf{F}, text description \textbf{C}, language \textbf{L}, and (optionally) the context \textbf{O}.
    \item For each function, generate \(k=10\) function-candidates based on the prompt. The response variety is maintained by the beam search technique. 
    \item Extract the functions in C or C++ from the model's predictions\footnote{For instance, for the CodeLlama model, the generated code snippets are usually marked with ``` symbol. Therefore, some additional post-processing is required.}.
\end{enumerate}

The set of function-candidates can be assessed in two main ways: qualitatively (read the code) and quantitatively (metrics). 

We considered the following metrics to evaluate models using our benchmark: Pass@k, Exact Match, and Edit Similarity. Although the metrics list is not exhaustive and can be extended, the following metrics are commonly used.

\begin{itemize}
    \item \textbf{pass@k} measures the functional correctness of a generated function-candidate, i.e. all tests from the repository should pass after the function-candidate replaced an original function,
    \item \textbf{Exact Match (EM)} indicates what proportion of examples has generated code being identical to the original code (this metric helps detect LLM's memorization),
    \item \textbf{Edit Similarity (ES)} indicates an average similarity score based on Levenshtein Distance between the generated and original code.
\end{itemize}

We computed the metrics in the following steps. First, for each repository, we prepared a separate environment by doing the following:
\begin{enumerate}
    \item loading repository;
    \item building a docker image with the repository environment: libraries, packages, etc;
    \item building project inside the docker image;
    \item running test suites and making sure all tests are passed.
\end{enumerate}

\begin{figure}
    \vspace{-5pt}
    \centering
    \includegraphics[width=0.40\textwidth]{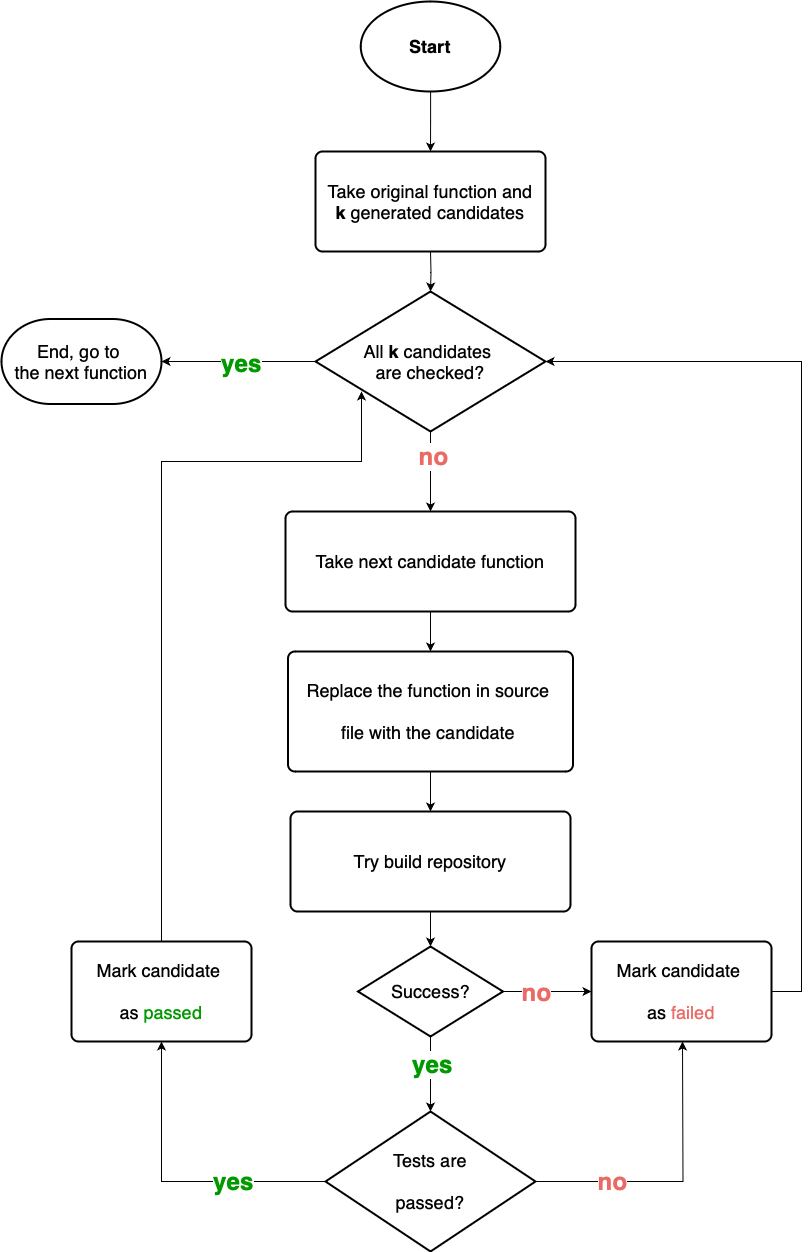}
    \caption{Overview of the pipeline for testing generated functions and \textit{pass@k} metric evaluation}
    \label{fig:testing}
    \vspace{-15pt}
\end{figure}

Second, inside each environment, we ran a script to calculate the pass@k metric. 
This procedure is described for an arbitrary function in the benchmark in Figure \ref{fig:testing}. 

Due to large repository sizes, the pass@k score calculation took a long time. This pipeline instantiated one container per generated function, and still, the evaluation took up to several hours on a single machine.

\begin{figure}
    \vspace{-10pt}
    \centering
    \includegraphics[width=0.5\textwidth]{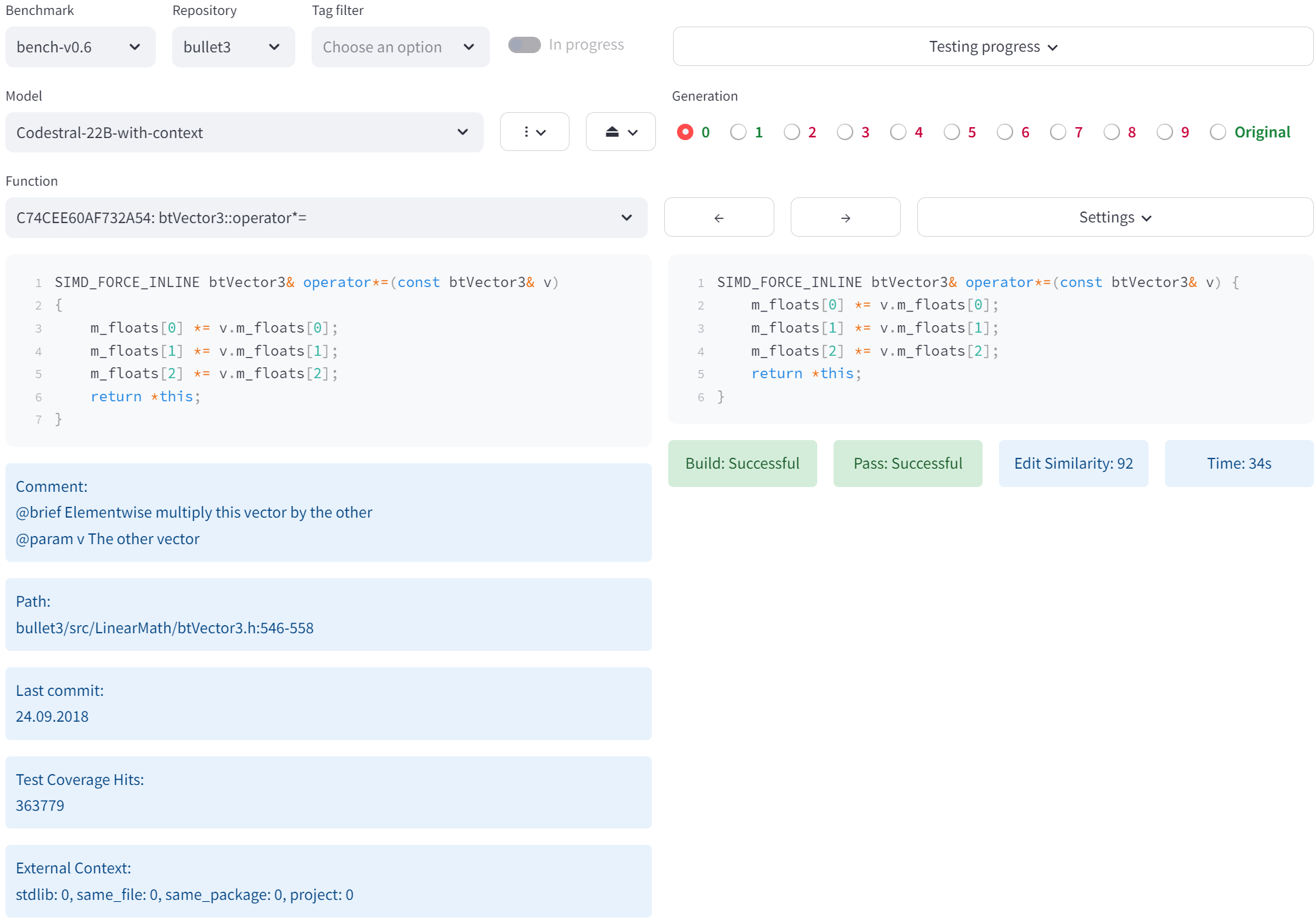}
    \caption{Visualization tool shows an original code snippet on the left and a series of ten generated snippets on the right.}
    \label{fig:vistool}
    \vspace{-20pt}
\end{figure}

Another approach to evaluation was qualitative, that is visualization of generated functions.
For visual analysis, we implemented a tool using Streamlit (Figure \ref{fig:vistool}). This tool allowed us to select a particular function from model generations and compare it with the original one from the repository. We could also see details about the snippet, e.g. docstring, its file path and span, last commit date, and test coverage hits. The tool also showed build and test statuses for every generation. A separate pop-up window was responsible for demonstrating testing progress including Pass@k, Exact Match, Edit Similarity, and time spent on testing.

\section{Experimental Setup}
\label{sec:exp-setup}
LLM generation quality depends on the quality of the prompt and, especially, on the context. The context can be produced in multiple ways, e.g. with few-shot examples, or retrieval of similar functions from an existing database. 
Therefore, we organized a series of experiments to assess the consistency and to check the benchmark sanity. These experiments were guided by a natural research question (`What is a baseline quality with both open source and proprietary LLM?'), which can be answered using the benchmark. 
In so doing, we aimed to check if the YABLoCo-based evaluation makes sense, is consistent, and correlates with previous research.



For the baseline models, we picked models CodeLlama-13B-Instruct\footnote{CodeLlama checkpoint \url{https://huggingface.co/codellama/CodeLlama-13b-Instruct-hf}}, DeepSeekCoder33B-Instruct\footnote{DeepSeekCoder checkpoint \url{https://huggingface.co/deepseek-ai/deepseek-coder-33b-instruct}} and  GPT-4. CodeLlama-13B-Instruct and DeepSeekCoder33B-Instruct served our needs since they (a) have quite good code generation performance based on various benchmarks, (b) were open source, and (c) could be run on moderate resources. The GPT-4 model was used for comparison to a closed-source solution. 

To set a baseline, we ran the inference with no context. We used the following template for prompt without context, combining the function with signature $[[F]]$, text description $[[C]]$, and language $[[L]]$ as follows.  
\begin{verbatim}    
    <s>[INST]<<SYS>>
    Generate code on L.
    <</SYS>>
    Function name: F. 
    Function description: C[/INST]}.
\end{verbatim}


The ultimate goal for the new dataset was leveraging the context of a large repository with LLM for code. A common way to use the context is an application of retrieval-augmented generation. This application is an active and wide area of research and cannot be investigated in the current paper. However, we provide results of an `oracle' context usage to show that the benchmark is suitable for similar experiments. 

By the `oracle' context we mean all the functions in the repository alongside docstrings and function bodies that are invoked in the target function body. Passing the oracle context should help a model to generate a function body that is similar to the original one. 
We extracted the `oracle' context as an intermediate step in call graph construction.

We then combined a prompt for the sample function with 
name $[[F]]$, text description $[[C]]$, and language $[[L]]$ as above. The `oracle' context $[[O]]$ was added as follows: 
\begin{verbatim}
    <s>[INST]<<SYS>>
    Generate code on [[L]].
    <</SYS>>
    Function name: [[F]]. 
    Function description: [[C]].
    Use this context: [[O]].[/INST]
\end{verbatim}

\section{Experimental Results with Baselines}

This section presents the results of baseline models evaluation on YABLoCo. 

We chose pass@k as the most representative metric; however, we also showed EM and ES metrics for the sake of validity. Tables \ref{tab:metrics-no_context-CLlama}-\ref{tab:metrcs-context-level} present more detailed results.

\subsection{In-context learning evaluation of the baseline LLMs}
Tables \ref{tab:metrics-no_context-CLlama}, \ref{tab:metrics-no_context-GPT} and \ref{tab:metrics-no_context-DSC} show the results of the experiments with in-context learning for the baseline LLMs. In all cases, prompts did not contain any repository context. 
Note that EM and ES metrics for the \textit{bullet3} repository have high values. 

\begin{table}[h!]
    \vspace{-10pt}
    \centering
    \caption{Baseline model \#1: CodeLlama-13B-Instruct without repository context}
    \begin{tabular}{|l|c|c|c|}
    \hline
         \textbf{Repository} & \textbf{Pass@10} & \textbf{Exact Match} & \textbf{Edit Similarity} \\ \hline
         bullet3    & 31.8  & 0.9 & 47.07 \\
         openssl    & 13.3  & 0.7 & 49.81 \\
         redis      & 20.0  & 0.0 & 50.58 \\
         llvm       & 13.4  & 0.0 & 45.96 \\ \hline
         overall    & 17.29 & 0.3 & 48.55 \\ \hline
         
    \end{tabular}
    \vspace{-10pt}
    
    \label{tab:metrics-no_context-CLlama}
\end{table}

\begin{table}[h!]
    \vspace{-10pt}
    \centering
    \caption{Baseline model \#2: GPT-4 without repository context}
    \begin{tabular}{|l|c|c|c|}
    \hline
         \textbf{Repository} & \textbf{Pass@10} & \textbf{Exact Match} & \textbf{Edit Similarity} \\ \hline
         bullet3    & 72.7  &  16.4 & 67.62 \\
         openssl    & 28.3  &  0.0  & 55.05 \\
         redis      & 29.2  &  0.9  & 54.52 \\
         llvm       & 19.4  &  1.2  & 51.32 \\ \hline
         overall    & 30.4  &  2.3  & 55.01 \\ \hline
    \end{tabular}
    \vspace{-10pt}
    \label{tab:metrics-no_context-GPT}
\end{table}

\begin{table}[h!]
    \centering
    \caption{Baseline model  \#3:  DeepSeekCoder-33B-Instruct without repository context}
    \begin{tabular}{|l|c|c|c|}
    \hline
         \textbf{Repository} & \textbf{Pass@10} & \textbf{Exact Match} & \textbf{Edit Similarity} \\ \hline
         bullet3    & 60.87  & 7.4 & 50.02 \\
         openssl    & 19.35  & 0.0 & 46.01 \\
         redis      & 20.59  & 0.0 & 45.7 \\
         llvm       & 14.29  & 0.0 & 37.95 \\ \hline
         Avg.    & 22.42 & 0.8 & 43.8 \\ \hline
         
    \end{tabular}
    \vspace{-10pt}
    \label{tab:metrics-no_context-DSC}
\end{table}

\subsection{Usage of oracle context}
We investigated the impact of the `oracle' context in prompt on two baselines.
Tables \ref{tab:metrics-oracle-codellama} and \ref{tab:metrics-oprinal-docs-deepseek} present metrics for CodeLlama-13B-Instruct and DeepSeekCoder-33B models.

\begin{table}[h!]
    \centering
    \caption{CodeLlama-13B-Instruct with the `oracle' context}
    \begin{tabular}{|l|c|c|c|}
    \hline
         \textbf{Repository} & \textbf{Pass@10} & \textbf{Exact Match} & \textbf{Edit Similarity}  \\ \hline
         bullet3    & 50.0  & 0.5 & 47.73 \\
         openssl    & 29.31 & 0.9 & 50.54 \\
         redis      & 30.77 & 0.2 & 51.85 \\
         llvm       & 22.73 & 1.1 & 48.38 \\ \hline
         overall    & 29.38 & 0.4 & 49.98 \\ \hline
    \end{tabular}
    \vspace{-10pt}
    
    \label{tab:metrics-oracle-codellama}
\end{table}

\begin{table}[h!]
    \centering
    \caption{DeepSeekCoder-33B with the `oracle' context}
    \begin{tabular}{|l|c|c|c|}
    \hline
         \textbf{Repository} & \textbf{Pass@10} & \textbf{Exact Match} & \textbf{Edit Similarity}  \\ \hline
         bullet3    & 81.82 & 7.7  & 51.22 \\
         openssl    & 33.9  & 0.2  & 52.99 \\
         redis      & 27.69 & 0.0  & 49.8 \\
         llvm       & 31.34 & 1.9  & 47.33 \\ \hline
         overall    & 36.15 & 1.5 & 50.05 \\ \hline
    \end{tabular}
    \vspace{-10pt}
    \label{tab:metrics-oprinal-docs-deepseek}
\end{table}

\begin{table}[h!]
    \centering
    \caption{Pass@10 results for different models across context levels. Models with '*' were augmented with oracle context from the corresponding project}
    \begin{tabular}{|l|c|c|c|c|c|}
    \hline
         \textbf{Model}  & \textbf{none} & \textbf{stdlib} & \textbf{file} & \textbf{package} & \textbf{project} \\ \hline
         CodeLlama-13B-Ins.  & 16.28 & 27.27 & 21.74 & 10.91 & 13.51\\
         DeepSeekCoder-33B  & 24.44 & 34.29 & 19.15 & 21.43 & 15.0\\
         GPT-4  & \textbf{32.56} & 33.33 & 30.43 & 34.55 & 18.92 \\
         CodeLlama-13B-Ins.*  & 13.95 & \textbf{36.36} & \textbf{51.11} & 24.07 & 25.0 \\
         DeepSeekCoder-33B*  & 20.93 & \textbf{36.36} & 41.3 & \textbf{40.0} & \textbf{41.67} \\
         \hline
    \end{tabular}
    \vspace{-10pt}
    \label{tab:metrcs-context-level}
\end{table}


Clearly, the addition of the `oracle' context improved the pass@k metric for both models (see the reference results in Tables \ref{tab:metrics-oracle-codellama} and \ref{tab:metrics-no_context-DSC}) as we compared only the settings where the `oracle' context was available to the LLMs. 

Table \ref{tab:metrcs-context-level} presents context level results. Clearly, providing the `oracle' context improved the baseline models for all context level groups, except for the `none` context level, where it hurt performance. For this `none` context level, GPT4 without repository context showed the highest score.

\section{Discussion of Results}

The above experimental results show how well the YABLoCo benchmark can measure baseline performance and possible improvements. The results from Tables \ref{tab:metrics-no_context-CLlama} and \ref{tab:metrics-no_context-GPT} imply that the baseline performance is quite moderate (20–30\% pass@10). CoderEval\cite{CoderEval} reports similar results. Moreover, GPT-4 outperforms CodeLlama on the benchmark by pass@k metric. Although not surprising, this result confirms the ability of YABLoCo to capture the quality of tested LLMs.

Next, as YABLoCo aims at measuring repository context code generation, we can refer to the results from the Tables \ref{tab:metrics-no_context-CLlama} and \ref{tab:metrics-oracle-codellama} to compare the baseline performance of CodeLlama with repository context. Here, we observe a significant improvement as CodeLlama supported by the `oracle' context reaches the performance of GPT-4. 


This improvement stimulates further research when different types of contexts can be mixed in the prompt to achieve even better performance. 
Note that the YABLoCo benchmark was not saturated in all our ICL experiments leaving room for improvement by leveraging different types of contexts or other code generation models.

During the investigation of relatively high `bullet3' results, we found that several functions from the benchmark had duplicate implementations that differed only in a type or a data structure presented in signatures. We hypothesize that some of these duplicates could survive deduplication and occur in a pre-training set leading to memorization and higher results.

\section{Related work}
\label{sec:relatedwork}

Measuring the performance is a crucial issue for comparison, empirical modeling, and improvement of code generation quality. To deal with this problem, many studies have released few benchmarks to allow a fair comparison between models. Recently, 
the question of repository-level code generation has attracted a lot of attention. This led to novel benchmarks for performance evaluation. We review the benchmarks for code generation that include cross-file or repository context. Their characteristics are summarized in Table \ref{tab:review-bench}.

\begin{table*}[t!]
\centering
\caption{Comparison of context benchmarks for code completion or generation. In the `Level' column `F' denotes Function, `S' denotes Statement}
\label{tab:review-bench}
\begin{tabular}{| l | l | l | l | l | l |}
\hline
\textbf{Benchmark} & \textbf{Level} & \textbf{Context} & \textbf{Languages} & \textbf{Size} & \textbf{Metric} \\
\hline
CoderEval \cite{CoderEval}   & F & In-file / Class / Repo & Python / Java & 230+230 &  Pass@k \\
\hline
RealCode\_eval\footnote{\url{https://github.com/NLP-Core-Team/RealCode\_eval}}   & F & In-file / Repo & Python & 219 functions & Pass@k \\
\hline
RepoBench \cite{RepoBench}  & S & In-file / Cross-file & Python / Java & 5K-18K (per lang)  & EM, ES \\
\hline
CrossCodeEval \cite{CrossCodeEval}  & S & In-file / Cross-file & Python / Java / TypeScript / C\# & 10K examples & EM, ES \\
\hline
Stack-Repo \cite{RepoFusion} & S & Repo & Java & 200 repos &  EM \\
\hline
RepoEval \cite{RepoCoder} & S & Repo & Python & 1600 examples & EM, ES \\
\hline
CatCoder \cite{pan2024enhancing} & F & Repo & Java / Rust & 199+90 examples & Pass@k \\
\hline
EvoCodeBench \cite{Li2024EvoCodeBenchAE} & F & Repo &  Python & 275 examples & Pass@k, Recall@k \\
\hline
\hline
YABLoCo & F & Repo & C/C++ &  215 functions & Pass@k \\
\hline
\end{tabular}
    \vspace{-10pt}
\end{table*}

\subsection{Benchmarks for Repository-level Code generation}
In RepoBench\cite{RepoBench}, Liu et al. focus on the code auto-completion task and collect a corresponding dataset in Python and Java. 
These authors develop retrieval and code completion mechanisms as well as their combination and study effect of cross-file and in-file context. Finally, Liu et al. highlight the importance of dealing with longer and more complex contexts. However, RepoBench lacks unit tests, which disallows computing the pass@k metric.
CrossCodeEval\cite{CrossCodeEval} addresses the same problem as RepoBench which is to introduce cross-file context into the code generation setting. Ding et al. \cite{CrossCodeEval} show that providing relevant context results in clear improvements for code LLMs. CrossCodeEval consists of 10K examples in four languages: Python, Java, TypeScript, and C\#. Ding et al. use Exact Match and Edit Similarity as metrics. 
RealCode\_eval (\url{https://github.com/NLP-Core-Team/RealCode\_eval}) aimed to diversify the most popular benchmarks such as LeetCode and CodeForces based on competition by introducing code generation examples from real repositories. RealCode\_eval is a dataset of 219 Python functions from 22 GitHub repositories published in the summer 2023.

Zhang et al. \cite{RepoCoder} propose their RepoCoder platform to approach the code completion problem using different files. It simplifies the code completion at the repository level by incorporating a similarity-based extractor and a pre-trained language model for code into an iterative extract generation pipeline. Moreover, a proposed benchmark, RepoEval, consists of the latest and highest quality real-world repositories covering scripted strings, API calls, and function body completion. 
However, RepoCoder focuses on smaller repositories, as the testing within larger repositories is time-consuming. Zhang et al. select 373 functions containing 3 to 30 lines of code. 

Shrivastava et al. \cite{RepoFusion} demonstrate the benefits of training using a repository context. In their Stack-Repo \cite{RepoFusion}, the authors propose RepoFusion, a framework for training models to include repository context. Single-line code completion experiments show that models trained on repository context significantly outperform larger code models.
Ablation studies show performance can depend on how design choices such as context type, number of contexts, and context length. The Stack-Repo dataset has 200 Java repositories, enriched with three types of contexts, and uses Exact Match for evaluation. 


CoderEval \cite{CoderEval} is a benchmark with different levels of context for Python and Java. It contains 230 examples from 10 projects in Java and 230 examples from 43 projects in Python. 
A key feature of the CoderEval is that it supports code generation tasks at six levels of context-dependency: \textit{self-contained}, \textit{slib-runnable}, \textit{plib-runnable}, \textit{class-runnable}, \textit{file-runnable}, and \textit{project-runnable}. Here, context refers to code elements such as types, APIs, variables, and constants defined outside the generated function, but within dependent third-party libraries, the current class, file, or project. CoderEval was used to evaluate the performance of models that generate code beyond standalone functions. 
Yu et al. \cite{CoderEval} also evaluated the quality of text descriptions of functions and found that using the exact texts from comments as prompts for LLM is suboptimal; human-labeled docstrings are more appropriate. 
Yu et al. use only the \textit{oracle\_context}, that is the information which the original function depends on. Overall, the CoderEval employs elaborate methodology for data selection, filtering, curating, and evaluation.

To the best of our knowledge, the two most recent benchmarks for code generation evaluation are the EvoCodeBench by Li et al.\cite{Li2024EvoCodeBenchAE} and CatCoder by Pan et
al. \cite{pan2024enhancing}. In EvoCodeBench, Li et al. identify three key problems of code generation quality assessment: keep code distribution as close to real-world repositories as possible, use a robust metric (pass@k), and avoid possible data leakage. 
A key difference to our work is repository size. 
EvoCodeBench has only 2 out of 25 repositories with 100-200K LoC. Although we experimented with different languages, we have arrived at similar conclusions to those from the EvoCodeBench. In deed, in some cases context might be not necessary at all, and the `More contexts benefit code generation' statement is arguable. Our experiments showed that only a more correct or valuable context could increase the performance, not any kind of repository context.

Finally, the CatCoder by Pan et al. \cite{pan2024enhancing} comprises 199 code generation tasks in Java and 90 tasks in Rust. Pan et al. focus on statically typed programming languages. The authors investigate the influence of the type contexts and code retrieval on the code generation quality.

\subsection{Retrieval-Augmented Generation}
Repository-Level Prompt Generation for Large Language Models of Code \cite{repoprompt} and RepoFusion \cite{RepoFusion} use the repository structure through prepared prompts that include various context templates and then select the most suitable one using a classification model. In contrast, RAMBO \cite{rambo} and RepoHyper \cite{repohyper} require no manual tagging of context templates, but use static analysis of the repository by splitting the code into elementary parts. Specifically, RAMBO splits the repository into methods, fields, and classes for all files and then extracts the relevant context. RepoHyper follows a similar approach but additionally builds a graph of dependencies between entities that describes most of the relationships between functions, classes, and files. This information-rich graph is then used to extract the context in two stages: breadth-first traversal and further filtering using GNN.

RepoCoder \cite{RepoCoder} and RAMBO \cite{rambo} use iterative data extraction to improve code generation. RepoCoder solves the problem of code completion, while RAMBO solves a narrower problem: generating a method body.
RepoCoder uses the whole source file to retrieve context and generate code, while RAMBO only uses the method code. According to Bui et al. \cite{rambo}, the iterative application of the above approaches improves the quality of code generation.

Apart from adding a textual representation of the context into the prompt, other methods exist to incorporate relevant context into LLM. The first example of such methods is Fusion in Decoder (FiD) \cite{fid}. This method encodes the relevant contexts separately and concatenates them into a single vector. Another example is KG-FiD \cite{kgfid} that uses a similar approach combining it with graph neural networks on knowledge graphs. The final example is RepoFusion \cite{RepoFusion} which integrates context from the entire code repository into the decoder using FiD.

\section{Limitations}
\vspace*{-3pt}
\label{sec:limitations}
The first and major limitation of the benchmark is possible data leakage, because some benchmark examples might appear in pre-train data of novel LLMs. To assess the risk of memorization of code to generate we included the Exact Match metric. From our experiments, we could conclude that the percentage of exact copies in the generated code was neglectfully small on average; however, for the \textit{bullet3} repository that percentage was moderate. For obvious reasons, the pass@k correlates with the exact match. To mitigate this risk, we plan to obfuscate the signatures of the functions that are used in prompts.

The second limitation of the benchmark is the approach to testing. We used the existing tests from the repositories. However, such usage might be a suboptimal approach as more test coverage of the benchmark functions would give a more precise pass@k value. However, higher test coverage would only decrease the pass@k metric, which was quite low in our results. Therefore, writing new tests is considered time-consuming and unnecessary for our benchmark at the moment.

The third limitation is that we used the original docstrings in prompts to generate code. However, the docstrings texts need to improved. For example, in CoderEval\cite{CoderEval} programmers reformulated docstrings manually. In the best-case scenario, such improvement should be done by the experts who know the corresponding projects.  

The fourth limitation is that the scope of this project was limited to only C/C++ and four selected repositories. These languages were chosen for novelty since most of the benchmarks focus on Java and Python only. We only had four repositories since adding new repositories involved a lot of manual effort, including building and compiling projects, writing Docker files, and automating running tests. We also limited the potential for generalization by setting the size of the benchmark to 215 examples. The reason for such a limit was that running all tests for a single model already takes 10-15 hours on a mediocre CPU.

To address the first and the third limitations, we generated synthetic docstrings by GPT-4o for the functions in the benchmark. As an input to GPT-4o, we passed docstring, signature, and body and ask the model to describe it short and precise. The result of GPT-4o was considered as synthetic docstrings. 

We treated original docstrings similarly (see Section \ref{man_eval}), i.e. we assessed of the synthetic docstrings manually. More than half of these docstrings were matching meaning with the original docstrings, and almost all of the synthetic docstrings were considered enough for generation functions in the benchmark. In addition, we investigated the impact of synthetic docstings using the DeepSeekCoder-33B model. The metrics for generated code with synthetic and original docstrings are presented in Tables \ref{tab:metrics-gpt-docs-deepseek} and \ref{tab:metrics-oprinal-docs-deepseek}, respectively.
 
\begin{table}[h!]
    \vspace{-10pt}
    \centering
    \caption{DeepSeekCoder-33B on synthetic docstrings with the `oracle'}
    \begin{tabular}{|l|l|c|c|c|}
    \hline
         Repository & Pass@k & Exact Match & Edit Similarity  \\ \hline
         bullet3    & 77.27 & 6.4 & 53.42 \\
         openssl    & 32.2  & 0.5 & 53.7 \\
         redis      & 30.77 & 1.5 & 40.91 \\
         llvm       & 26.87 & 2.4 & 49.49 \\ \hline
         overall    & 34.74 & 2.0 & 51.44 \\ \hline
    \end{tabular}
    \vspace{-5pt}
    \label{tab:metrics-gpt-docs-deepseek}
\end{table}

As a result, pass@k dropped 1 to 4 points. We suggest to use the synthetic docstrings created by the GPT-4o as a "hard" version of the benchmark. 

\section{Conclusion and Future work}
This paper contributes a novel benchmark for code generation evaluation in large repositories. This benchmark consists of four repositories in C/C++ and an evaluation pipeline for efficient metric calculation. The benchmark has a few unique features, including the largest context sizes 400K-2M LoC, and five categories of context dependency for functions. Therefore, we claim that the YABLoCo benchmark will help address a practically relevant gap, that is how context from a single large repository can be used in generative language models for code. 

To demonstrate the consistency of the benchmark, we ran a series of experiments with multiple LLMs for code as to discover possible improvements that leverage repository context. These experiments confirmed that the benchmark was not easy to solve even in the case of advanced and proprietary pre-trained models, which could have seen the repositories. The dataset itself might be a useful test bed for other kinds of investigations and case studies of generative AI applications in software engineering. 
\section*{Acknowledgements}
All the authors affiliated with the Innopolis University were supported by the Research Center of the Artificial Intelligence Institute of Innopolis University.

\bibliographystyle{plain}
\bibliography{main}





\end{document}